\documentclass[sigconf]{acmart}

\AtBeginDocument{%
  \providecommand\BibTeX{{%
    \normalfont B\kern-0.5em{\scshape i\kern-0.25em b}\kern-0.8em\TeX}}}

\acmConference[arxiv]{ArXiV}

\usepackage{booktabs} 
\usepackage{graphicx}
\usepackage{amsthm}
\usepackage{array}
\usepackage{multirow}
\usepackage{xcolor}

\newcolumntype{P}[1]{>{\centering\arraybackslash}p{#1}}
\usepackage{float}
\usepackage[ruled]{algorithm2e} 

\SetAlFnt{\small}
\SetAlCapFnt{\small}
\SetAlCapNameFnt{\small}
\SetAlCapHSkip{0pt}
\usepackage{comment}
\usepackage{todonotes}

\newcommand{\ejain}[1]{ \color{red} EJ: #1 \color{black}}

\begin{document}

\title{In-vehicle alertness monitoring for older adults}

    \author{Heng Yao}
    \affiliation{%
      \institution{\textit{University of Florida}}
      \country{USA}
    }
    \email{hengyao@ufl.edu}
    
    \author{Sanaz Motamedi}
    \affiliation{%
      \institution{\textit{University of Florida}}
      \country{USA}
    }
    \email{smotamedi@ufl.edu}
    
    \author{Wayne C.W. Giang}
    \affiliation{%
      \institution{\textit{University of Florida}}
      \country{USA}
    }
    \email{wayne.giang@ise.ufl.edu}
    
    \author{Alexandra Kondyli}
    \affiliation{%
      \institution{\textit{University of Kansas}}
      \country{USA}
    }
    \email{akondyli@ku.edu}
    
    \author{Eakta Jain}
    \affiliation{%
      \institution{\textit{University of Florida}}
      \country{USA}
    }
    \email{ejain@cise.ufl.edu}

\renewcommand{\shortauthors}{Yao, Motamedi, Giang, Kondyli and Jain.}

\begin{abstract}
 
Alertness monitoring in the context of driving improves safety and saves lives. Computer vision based alertness monitoring is an active area of research.
However, the algorithms and datasets that exist for alertness monitoring are primarily aimed at younger adults (18-50 years old). 
We present a system for in-vehicle alertness monitoring for older adults.
Through a design study, we ascertained the variables and parameters that are suitable for older adults traveling independently in Level 5 vehicles. We implemented a prototype traveler monitoring system and evaluated the alertness detection algorithm on ten older adults (70 years and older). We report on the system design and implementation at a level of detail that is suitable for the beginning researcher or practitioner.
Our study suggests that dataset development is the foremost challenge for developing alertness monitoring systems targeted at older adults. 
This study is the first of its kind for a hitherto under-studied population and has implications for future work on algorithm development and system design through participatory methods.  
\end{abstract}




\maketitle

\section{Introduction}
The ability to travel independently within a city is a major factor in quality of life for older adults. Older adults, especially those with age-related cognitive disabilities, may stop driving due to deficits in reaction time, memory, and attention. Connected and automated vehicles (CAVs) prolong mobility and safety for this population. CAVs are those vehicles where at least some aspect of a safety-critical control function (e.g., steering, throttle, or braking) occurs without direct driver input~\cite{national2013preliminary}. CAVs may be autonomous (i.e., use only vehicle sensors) or may be connected (i.e., use communications systems such as connected vehicle technology, in which cars and roadside infrastructure communicate wirelessly). Motivated by technological innovations in this direction, The United States Department of Transportation recently announced an inclusive design challenge. This challenge invited proposals for highly automated vehicle designs that centered persons with disabilities. 

In 2020, there were an estimated 38 million people aged 70 and older living in the United States, representing about 11 percent of the population \cite{federal-highway-administration-2022}. Based on data reported by states to the Federal Highway Administration, there were approximately 31 million licensed drivers 70 and older in 2020 \cite{federal-highway-administration-2022}. According to the U.S. Census Bureau, the population 70 and older is projected to increase to 53 million by 2030~\cite{us2017national}. 

Cognitive impairment is a common problem for older adults~\cite{rashid2012cognitive, reitz2010use}. Cognitive decline associated with age~\cite{tricco2012use, rashid2012cognitive, ataollahi2014does}, has an occurrence rate of approximately 21.5–71.3 per 1,000 person-years in seniors~\cite{tricco2012use}. Mild cognitive impairment (MCI) rates range from 3\% to as high as 42\% in population studies~\cite{tricco2012use}, and from 6\% to 85\% in clinical settings~\cite{pinto2009mild, petersen1999mild}. The MCI conversion rate to dementia is about 10\% per year~\cite{decarli2003mild, pinto2009mild, rc2001practice, rashid2012cognitive, bruscoli2004mci, petersen2004mild, panza2005current}, which is increased to 80\%–90\% after approximately 6 years~\cite{busse2006progression}. It is estimated that a new case of dementia is added each 7 seconds~\cite{rashid2012cognitive}. The prevalence of dementia in the elderly population is between 1\% to 2\% per year~\cite{odawara2012cautious, smith1996definition}. It is forecast that the number of cases in the developing world will increase by 100\% between 2001 and 2040~\cite{rashid2012cognitive}. It seems that the rate of dementia will increase from 9.4\% in 2000 to 23.5\% by 2050 in the population over 60 years of age~\cite{nizamuddin2000population}.

These statistics motivate the ultimate goal of our project: to design and characterize methods for in-vehicle alertness monitoring for older adults with mild cognitive impairment (MCI). As this population experiences reduced alertness and memory, difficulty in information processing, memory loss, disorientation, and difficulty in decision-making, it is important to monitor their alertness level to ensure their safety and improve their mobility experience~\cite{parikh2016impact, gold2012examination}. This paper reports on our design process and algorithmic implementations for an in-vehicle alertness monitoring system for older adults.

Driving automation technologies have been classified in a six-level system by SAE International \cite{sae-international-2021}. Our focus is on personal vehicles that will exhibit SAE Level 5 automation \cite{sae-international-2021}. The level 5 vehicles perform the dynamic driving task independently of the driver, under all conditions and in different scenarios. Fully autonomous cars are undergoing testing in several pockets of the world, but none are yet available to the general public \cite{novakazi2020perception}. 
Designing for independent traveling in an SAE Level 5 vehicle creates a scenario where attentiveness monitoring is distinct from driver inattention, as the traveler is not expected to attend to the driving tasks or even to take over. The in-vehicle monitoring system that we designed can be extended to incorporate eye movements, situational awareness, etc., for vehicles that have lower levels of automation. 

Our contributions are as follows:
\begin{enumerate}
    \item We develop an alertness monitoring system for a previously under-studied population. We evaluate the performance of our system and algorithm on the older adults.
    
    \item Our algorithm is based on four features drawn from prior literature: eye aspect ratio (EAR); mouth aspect ratio (MAR); MAR/EAR (MOE); pupil circularity. We are the first to report on the performance of these features on a large publicly available dataset (UTA Real-Life Drowsiness Dataset (UTA-RLDD) \cite{ghoddoosian2019realistic}.
    
    \item We present the design decisions behind a practical implementation of such a system in a prototype Level 5 vehicle. These design decisions range from screen and camera placement (hardware), to feedback about where the head is placed (user experience), to algorithmic implementation. We target our system design report to the beginning researcher or practitioner. 
\end{enumerate}

\section{Related Work}
Our work builds on prior research that has used facial landmark features to classify a person's alertness level. Our review background literature considered alertness, attentiveness, drowsiness and sleepiness to be part of the same goal, namely, improving safety, and thus we consider these terms interchangeable.

\subsection{Features used to detect alertness levels}
Multiple approaches were used in order to detect drivers’ attentiveness using sensorial, physiological, behavioral, visual, and environmental modalities. Early methods considered the steering motion of vehicles as an indicator of drowsiness~\cite{sahayadhas2012detecting}. More recent approaches used face monitoring and tracking to detect the alertness levels of drivers. These methods relied on extracting facial features related to yawning and eye closure \cite{park2016driver, reddy2017real, smith2000monitoring}. Additionally, approaches that measured physiological signals such as brain waves and heart rates were introduced in order to correlate them to states of drowsiness~\cite{kokonozi2008study}.

Several studies have been conducted on the state of the driver's eye based on a video surveillance system to detect the eyes and calculate their frequency of blinking to check the driver's fatigue level~\cite{massoz2018multi, mandal2016towards, al2013eye, kurylyak2012detection}. Some studies have used a cascade of classifiers to detect the ocular area's rapid detection~\cite{kurylyak2012detection}. A Haar-like descriptor and an AdaBoost classification algorithm have been widely used for face and eye-tracking by using Percent Eye closure (PERCLOS) to evaluate driver tiredness~\cite{lienhart2002extended, freund1996experiments}. PERCLOS evaluates the proportion of the total time that a drivers' eyelids are $\ge$80\% closed and reflects a slow closing of the eyelids rather than a standard blink \cite{wierwille1994research}.

Wahlstrom et al. developed a gaze detection system to determine the activity of drivers in real-time in order to send a warning in states of drowsiness~\cite{wahlstrom2003vision}. Gundgurti et al. extracted geometric features of the mouth in addition to features related to head movements to detect drowsiness effectively. However their method suffered in situations involving poor illumination and variations in the skin color~\cite{gundgurti2013experimental}. Sigari et al. developed an efficient adaptive fuzzy system that derived features from the eyes and the face regions such as eye closure rates and head orientation to estimate the drivers’ fatigue level using video sequences captured in laboratory and real conditions~\cite{sigari2013driver}. Rahman et al. proposed a progressive haptic alerting system which detected the eyes state using Local Binary Pattern and warned the drivers using a vibration-like alert~\cite{rahman2011novel}. Jo et al. introduced a system to separate between drowsiness and distraction using the driver’s face direction~\cite{jo2011vision}. The system utilized template matching and appearance-based features to detect and validate the location of the eyes using a dataset collected from 22 subjects~\cite{jo2011vision}. Taken together, these approaches faced challenges in practical adoption due to errors resulting from signal disambiguation, loss of facial tracking arising from sudden movements, and the invasiveness of signal extraction using contact-based sensors.

The detection of facial features (also called landmarks) is an essential part of many face monitoring systems. Eyes, as one of the most salient facial features reflecting individuals’ affective states and focus of attention~\cite{song2014eyes}, have become one of the most remarkable information sources in the face. Eye tracking serves as the first step in order to get glance behaviour, which is of most interest because it is a good indicator of the direction of the driver’s attention~\cite{devi2008driver}. Glance behaviour can be used to detect both visual and cognitive distraction~\cite{fernandez2016driver}. It has also been used by many studies as an indicator of distraction while driving and has been evaluated in numerous ways~\cite{fernandez2016driver}. Therefore, both eye detection and tracking form the basis for further analysis to get glance behaviour, which can be used for both cognitive and visual distraction.


\subsection{Face and Facial Landmarks Detection}
A typical face processing scheme in alertness monitoring systems involves the following steps: 
\begin{itemize}
    \item \textit{Face detection and head tracking}: In many cases a face detection algorithm is used as a face tracking one. In other cases, a face detection algorithm is used as an input for a more robust face tracking algorithm. When the tracking is lost, a face detection call is usually involved.
    \item \textit{Localization of facial features (e.g., eyes)}. Facial landmarks localization is usually performed, but it should be noted that, in some cases, no specific landmarks are localized. So, in such cases, estimation of specific cues are extracted based on anthropometric measures from both face and head.
\end{itemize}

In general, “detection” processes are machine-learning based classifications that classify between object or non-object images. For example, whether a picture has a face on it or not, and where the face is if it does. To classify, we need a classifier.

Our implementation is based on the Viola-Jones algorithm, which has made object detection practically feasible in real-world applications~\cite{vikram2017facial}. The Viola-Jones algorithm contains three main ideas that make it possible to build and run in real-time: the integral image, classifier learning with AdaBoost, and the attentional cascade structure~\cite{heydarzadeh2011efficient}. This framework is used to create state-of-the-art detectors (e.g., face detectors~\cite{fan2012system}), such as in the Opencv library. However, cascade detectors work well on frontal faces but sometimes, they fail to detect profile or partially occluded faces. Thus, we used the Histogram of Oriented Gradients (HOG), which is a feature descriptor used in computer vision and image processing for the purpose of object detection. This approach can be trained with fewer images and faster~\cite{surasak2018histogram}.

\subsection{Classification methods used to detect attentiveness levels}
The majority of methods for alertness detection in the last decade have utilized deep networks in one form or another.
For extracting the intermediate representation, algorithms consist of proprietary software~\cite{wang2016driver, liang2019prediction, franccois2016tests}, or a pre-trained CNN~\cite{shih2016mstn} such as the VGG-16~\cite{simonyan2014very}. For characterizing alertness, models consist of logistic regression~\cite{liang2019prediction}, artificial neural network (ANN)~\cite{wang2016driver}, support vector machine (SVM)~\cite{cristianini2000introduction}, hidden Markov model (HMM)~\cite{weng2016driver}, Multi-Timescale by CNN~\cite{massoz2018multi}, long-short term memory (LSTM) network smoothed by a temporal CNN~\cite{shih2016mstn}, or end-to-end 3D-CNN~\cite{huynh2016detection} are conducted. 
\section{Method}
We start with design considerations for an older adult alertness monitoring system, present our approach to determine key parameters, and finally describe the feature extraction and classification algorithms. 

\subsection{Design variables}
For the purpose of our use case, the Level 5 CAV was assumed to be personally owned, which meant that we expected that parameters could be personalized to one traveler. The vehicle was expected to be a mid-sized sedan or larger, such as an SUV or minivan. Because the design study is for SAE level 5 vehicles, we expected that travelers will sit in the vehicle in the same way as we sit in the front passenger seat in manually-driven vehicles of today. Therefore, no steering wheels or other vehicle control devices would be in front of the traveler. Instead, the main features of the vehicle cab would be a dashboard, with an interactive display that presents trip relevant information to the traveler. Sensors to monitor the traveler would be integrated into the dashboard or the seat, and finally feedback from the driver monitoring system would be shown on a display located on the dashboard.

In the list below, we discuss design variables specific to traveler alertness monitoring.

\begin{enumerate}
    \item Sensor
    \begin{enumerate}
        \item \textit{Type}: Alertness is a user state that can be operationalized in several ways, for example, as alertness versus sleepiness, or as situational attentiveness, or as focusing (foveating) on a key region of interest. For each of these definitions, there is a range of sensor types to achieve the goal. In this study, we chose visible light cameras as our primary traveler monitoring sensor for attentiveness
        We selected a standard webcam (Logitech C270 HD Webcam) as it is a low-cost sensor which can additionally be used for head pose detection and facial landmark tracking. 
        \item \textit{Placement}: The placement of the sensor within the vehicle must be done in such a way that it provides a natural advantage for recording, while at the same time minimally occluding the traveler's experience. We experimented with three configurations of camera placement and three distances from the traveler. 
    \end{enumerate}
    \item Traveler 
    \begin{enumerate}
        \item \textit{Expectations from the traveler}: This design variable refers to the extent to which the designer and developer can expect the user to cooperate with requests to align themselves to a tracking box, or submit to a calibration procedure, or sit still while a baseline is being established, or respond to audiovisual alerts, etc. 
        \item \textit{Characteristics of the user}: Older people have different physical characteristics compared to younger people, which can impact the outcome of a traveler monitoring system. For example, older adults will have wrinkles around the eye and are more likely to be wearing glasses. Other characteristics include the distribution of activities older adults perform, which can change the distribution of head and body poses the traveler monitoring system will receive compared to young users. Furthermore, there is some evidence that older adults with MCI may have different eye movements than healthy individuals~\cite{nie2020early, wilcockson2019abnormalities, molitor2015eye}. There are also likely to be changes in the distribution of activities, attentional focus, and lucidity for older adults with MCI.
    \end{enumerate}
    \item Data
    \begin{enumerate}
        \item \textit{Raw data}: This variable refers to the data that is recorded from the traveler. In our study, raw data consists of RGB video recorded at 24fps. No audio is required for alertness detection.
        \item \textit{Features}: This variable refers to how the raw data is processed, the time scale over which data is smoothed or otherwise preprocessed to reduce noise, and the extent to which features of interest are passed up the software stack. For example, video data may be processed to extract facial landmark points, which are sent to one computational sub-routine, and to extract the subset of pixels that comprise the eye region and mouth area, which are sent to a different computational sub-routine. Raw data may be secured on the vehicle while only specific features are permitted to be passed upstream to protect user privacy.
    \end{enumerate}
    \item Context: In addition to the features extracted from the primary sensor, other data may be used in the decision making process. For example, for alertness monitoring, the predictions from the video channel could be checked against a physiological sensor to rule out false alarms as a result of lighting and shadow. 
    \item Intelligence: Data recorded by one or more sensors needs to be fused for reliable predictions. In addition, this variable refers to the extent to which the underlying algorithm incorporates computational mechanisms to adapt to the primary traveler (see note on personally owned vehicle) or adopts a best fit (one-size-fits-all) approach.  
    \item Feedback: This design variable considers the extent to which the system is permitted to ask the traveler to reorient themselves for facial tracking, as well as the types of cues that the system will provide to the traveler, for example, visual and audio prompts to assess if the traveler simply dozed off or needs medical assistance can be provided by the vehicle. 
    \item Privacy: Because the traveler alertness monitoring system will record highly personal data, there is a trade-off between local processing versus cloud-based processing in terms of user privacy and algorithmic improvements which leverage the cloud-based computational back end. For example, a system where all facial landmark features are processed locally, and no face videos are stored, would address privacy concerns related with out of context use of the facial video stream.
    \item Security: Because the goal of an alertness monitoring system is to escalate the alerts if the traveler does not respond to a nudge, the design and implementation could create a novel attack space from the cybersecurity perspective. For example, a malicious adversary might gain access to the the alert code to get the vehicle to repeatedly call the caregiver, or perform a man-in-the-middle attack on the camera feed causing the vehicle suddenly come to a stop because it thinks the traveler needs medical attention.
\end{enumerate}

\subsection{Design of the alertness monitoring system}
\label{ref:system-design}

The purpose of the proposed system is to gather implicit cues from the older adult traveler to assess their degree of alertness, which allows the automated vehicle to monitor the passenger’s state and act accordingly. 
Several aspects of the traveler can be monitored and tracked such as: expressions, body positions, eye-tracking monitoring, physiological signals, etc. In our prototype, we have focused on alertness as our target population is susceptible to motion-induced sleepiness and distraction~\cite{ren2019cognitive, lin2016mental}. Discussions with the physicians on the team, the advisory board members, and driver rehabilitation specialists revealed that taking a nap may be considered entirely reasonable while on a long trip. However, it would be concerning for a short (<20 minutes) trip if the older adult with MCI traveling alone displayed signs of low alertness.

\begin{figure}[tb]
\centering
\includegraphics[width=\linewidth]{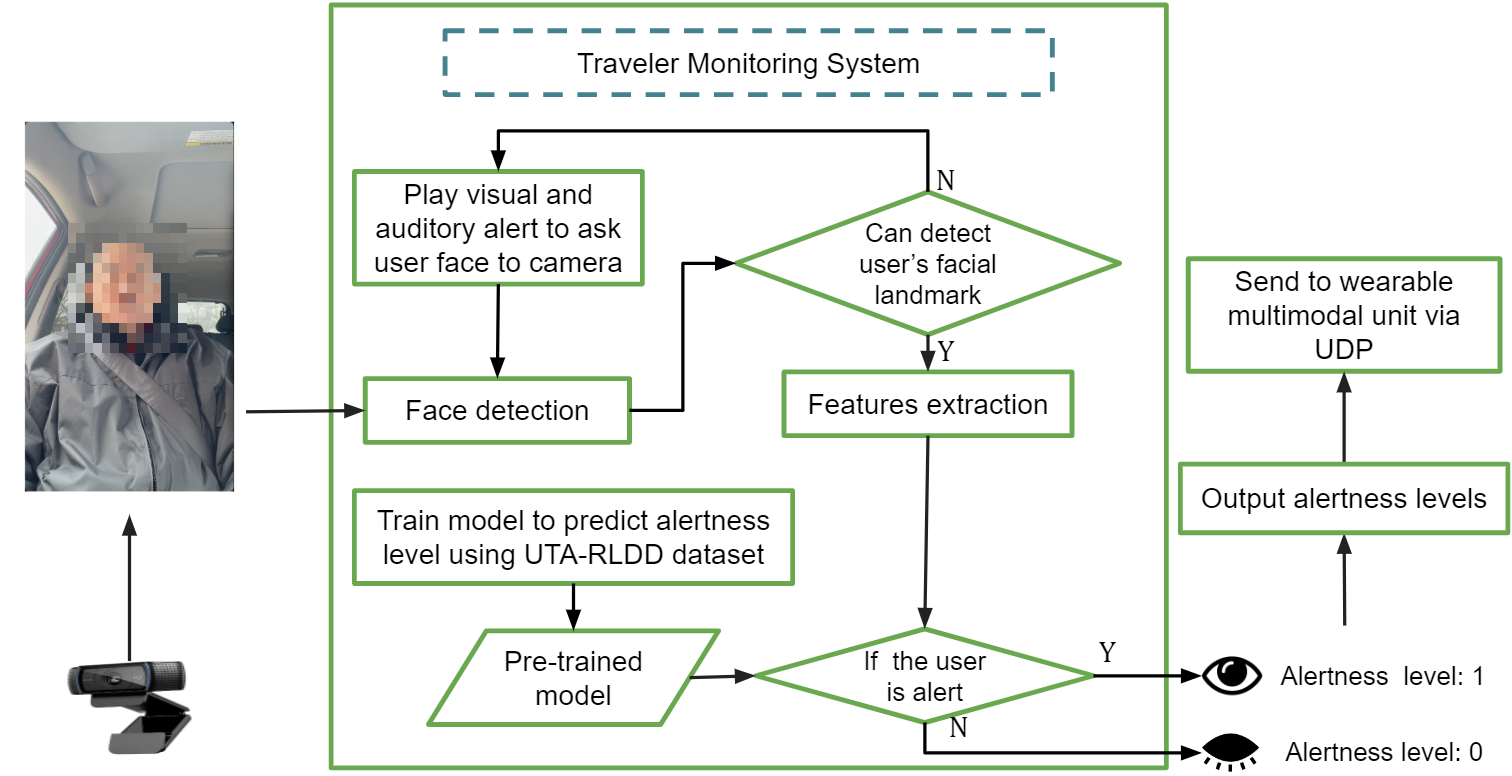}
\caption{Technical details of the computer vision-based traveler monitoring subsystem}
\label{fig:tmsdetail}
\end{figure}
\begin{figure}[tb]
\centering
\includegraphics[width=\linewidth]{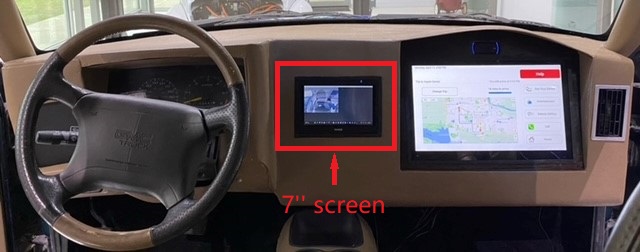}
\caption{7'' screen in the dashboard}
\label{fig:screenindash}
\end{figure}
\begin{figure}[tb]
\centering
\includegraphics[width=0.5\linewidth]{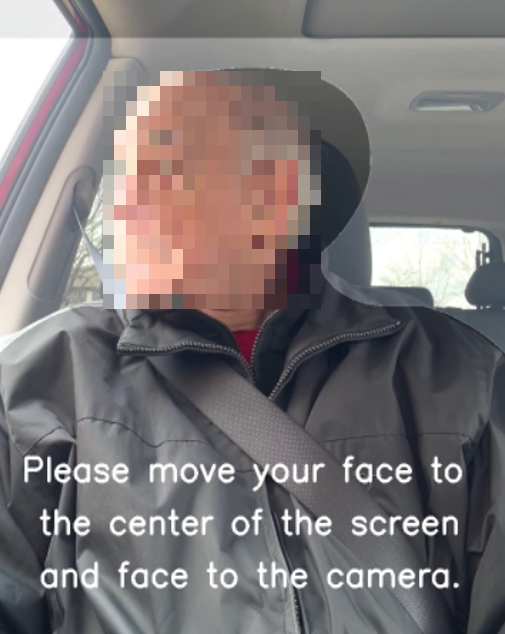}
\caption{Visual feedback to remind users to face the remote camera}
\label{fig:vfd}
\end{figure}

Figure~\ref{fig:tmsdetail} shows the system diagram. The prototype system uses computer vision libraries to monitor features associated with low alertness, such as eye aspect ratio, mouth aspect ratio, and pupil circularity, by using a remote camera embedded within the vehicle’s dashboard. These features are combined to predict an overall level of alertness of the traveler in real time. Each frame of the camera feed goes through the following processing steps: contrast enhancement, face detection, feature extraction, and alertness level prediction.  
Figure~\ref{fig:screenindash} illustrates the dashboard of the prototype vehicle. There is a small 7'' screen placed adjacent to a larger in-vehicle information system. The system plays a beep sound through a sound bar in the dashboard when it cannot detect the traveler’s face, to remind them to reposition back to the center of the screen. In addition to the auditory cue, the system provides visual cues as shown in Figure~\ref{fig:vfd}.

\begin{figure}[tb]
\centering
\includegraphics[width=\linewidth]{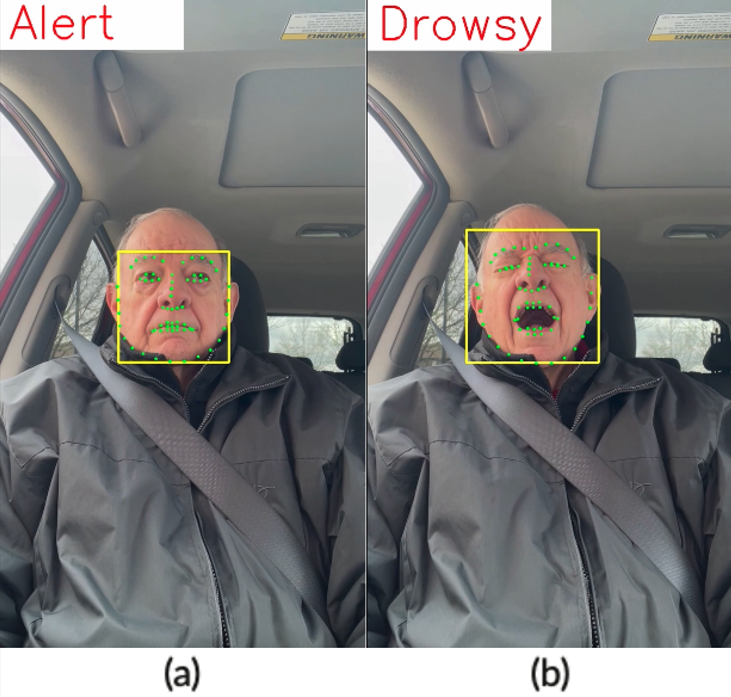}
\caption{Alertness level detection results (a) alert state (b) drowsy state}
\label{fig:detectionres}
\end{figure}

To account for individual differences in facial structure and the facial cues that indicate low alertness, the traveler monitoring system starts with a calibration step for each new traveler. During calibration, the traveler is asked to keep their face steady and centered on the screen in front of them for around 30 seconds. The algorithm assumes that these data contain exemplars of a high alertness state for this traveler. When the features deviate from this exemplar, the algorithm detects a loss in alertness level. The features are calculated at a frame-by-frame level. We average the features in ten consecutive frames. When the average features deviate from the exemplar capture in the calibration stage, the algorithm detects a loss in alertness. Examples of the detection result between alert and drowsy states are presented in Figure~\ref{fig:detectionres}. 

\subsection{Camera and screen placement}
\label{ref:camera-screen-placement}
Detection accuracy is highly dependent on the quality of input to the feature detection algorithm.
We tested different locations of the camera in a car aiming to capture the traveler's front face. A front face will enable us to extract features we need to predict the traveler's alertness level. Other than the camera, we also need to consider the location of visual displays in the dashboard, as the traveler will likely be looking at those displays.

\begin{figure}[tb]
\centering
\includegraphics[width=\linewidth]{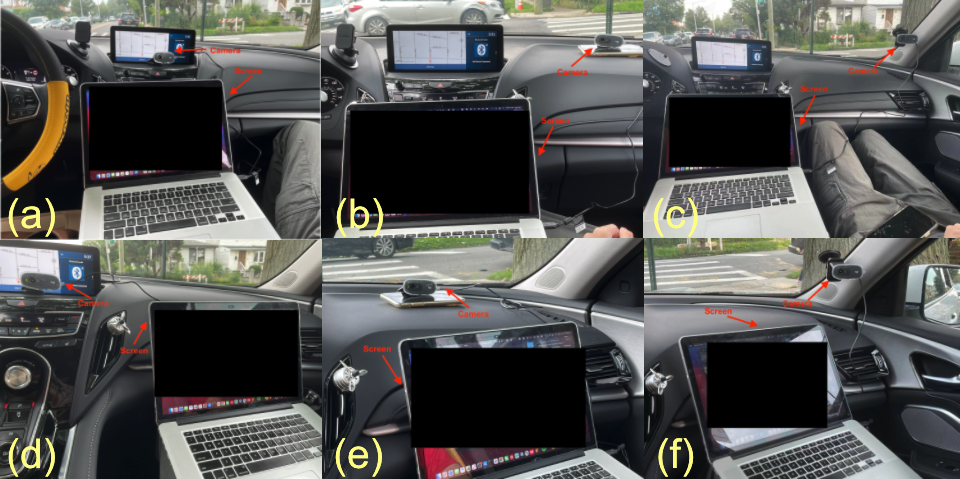}
\caption{Camera and screen placement}
\label{fig:camplace}
\end{figure}

To determine an optimal location of the camera for our system, we tested three different locations (In the middle of the dashboard - Figure \ref{fig:camplace}.(a) and Figure \ref{fig:camplace}.(d); In front of the traveler - Figure \ref{fig:camplace}.(b) and Figure \ref{fig:camplace}.(e); On the right side of the front windshield - Figure \ref{fig:camplace}.(c) and Figure \ref{fig:camplace}.(f)). We tested two locations for the screen (In the middle of the dashboard - Figure \ref{fig:camplace}.(a), Figure \ref{fig:camplace}.(b) and Figure \ref{fig:camplace}.(c); In front of the traveler - Figure \ref{fig:camplace}.(d), Figure \ref{fig:camplace}.(e) and Figure \ref{fig:camplace}.(f)). Table \ref{tab:fdr} shows the face detection rate in different camera and screen placement locations. Based on the percentage of successful face detections, the optimal location for both camera and screen is determined to be in front of the traveler.
 
To identify the optimal distance between users' eyes and the camera for our system, we tested four distances (44.3 inches, 39.2 inches, 30.5 inches, 28.2 inches). We recorded videos at these four distances and analyzed the facial landmark detection rate for each video. We recorded two videos for each distance while the user was wearing or not wearing glasses. We converted the video to image frames. We extracted five frames per second. For all the videos, we followed the following procedure to mimic a traveler's behaviors.
\begin{itemize}
    \item 0-10s: preparation (we removed this part before analyzing the face detection rate)
    \item 10-40s: looking ahead
    \item 40-70s: looking at the main display (the laptop)
    \item 70-100s: using a phone
    \item 100-130s: looking at the front left passenger seat
    \item 130-160s: looking ahead
    \item 160-190s: looking at the main display
    \item 190-220s: using a phone
    \item 220-250s: looking at the front left passenger seat
\end{itemize}

Table \ref{tab:fdrd} shows the face detection rate for all image frames of each video under different distances:

The result showed that the face detection rate increased when the camera was moved closer to the participant when they were not wearing glasses. When wearing glasses, the face detection rate generally increased as the camera was moved closer, but the accuracy dropped off at the closest distance (28.2 inches) suggesting that glasses may be interfering with face detection at this close range. Among the four tested distances, 30.5 inches appears to be the most optimal. Under this distance, we could detect facial landmarks for almost all frames (99.86\%) for both wearing and not wearing glasses conditions. Cameras placed at closer distances may be impacted by glasses and it the cameras may also be more obvious to users leading to annoyance.

\begin{table*}[]
\begin{tabular}{|cc|ccc|}
\hline
\multicolumn{2}{|c|}{\multirow{2}{*}{Face detection rate \%}}                                                                                                                                          & \multicolumn{3}{c|}{Camera location}                                                                                                                                                                                                                                            \\ \cline{3-5} 
\multicolumn{2}{|c|}{}                                                                                                                                                                                 & \multicolumn{1}{c|}{\begin{tabular}[c]{@{}c@{}}In the middle of\\ the dashboard\end{tabular}} & \multicolumn{1}{c|}{\begin{tabular}[c]{@{}c@{}}In front of\\ the traveler\end{tabular}} & \begin{tabular}[c]{@{}c@{}}On the right\\ side of the front\\ windshield\end{tabular} \\ \hline
\multicolumn{1}{|c|}{\multirow{2}{*}{\begin{tabular}[c]{@{}c@{}}Main display\\ location - below\\ the camera\end{tabular}}} & \begin{tabular}[c]{@{}c@{}}In the middle of\\ the dashboard\end{tabular} & \multicolumn{1}{c|}{91.34\%}                                                                  & \multicolumn{1}{c|}{99.11\%}                                                            & 71.36\%                                                                               \\ \cline{2-5} 
\multicolumn{1}{|c|}{}                                                                                                      & \begin{tabular}[c]{@{}c@{}}In front of\\ the traveler\end{tabular}       & \multicolumn{1}{c|}{79.01\%}                                                                  & \multicolumn{1}{c|}{96.08\%}                                                            & 85.64\%                                                                               \\ \hline
\end{tabular}
\caption{Face detection rate in different camera \& screen placement locations}
\label{tab:fdr}
\end{table*}
 
\begin{table*}[]
\begin{tabular}{|c|c|c|}
\hline
{\color[HTML]{201F1E} Inches (distance between eye and camera)} & Glasses (face detection rate) & No glasses (face detection rate) \\ \hline
44.3                                                            & 88.07\%                       & 99.61\%                          \\ \hline
39.2                                                            & 97.71\%                       & 99.65\%                          \\ \hline
30.5                                                            & 99.86\%                       & 99.86\%                          \\ \hline
28.2                                                            & 94.46\%                       & 100\%                            \\ \hline
\end{tabular}
\caption{Facial landmark detection rate in different camera \& screen placement distances}
\label{tab:fdrd}
\end{table*}

\subsection{Feature Extraction}
\label{sec:features}
We used dlib’s pre-trained facial landmark to detect a face area and extract features. We selected four features from the previous literature to train our classification model~\cite{sathasivam2020drowsiness, singh2018driver}. We also tested the performance of classifiers trained using different combinations of these features.

\subsubsection{Eye Aspect Ratio (EAR)}
\begin{figure}
\centering
\includegraphics[width=0.5\linewidth]{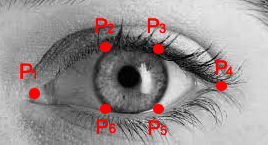}
\caption{Eye aspect ratio}
\label{fig:ear}
\end{figure}

EAR is the ratio of the length of the eyes to the width of the eyes. The length of the eyes is calculated by averaging over two distinct vertical lines across the eyes, as illustrated in the Figure~\ref{fig:ear} and Equation~\ref{eqn:EAR}. Based on literature~\cite{sathasivam2020drowsiness}, if an individual becomes drowsy, their average eye aspect ratio over successive frames starts to decline since their eyes started to be more closed or they were blinking faster.

\begin{equation} 
\label{eqn:EAR}
EAR=\frac{\left \| p_{2}-p_{6} \right \|+\left \| p_{3}-p_{5} \right \|}{2\left \| p_{1}-p_{4} \right \|}
\end{equation}

\subsubsection{Mouth Aspect Ratio (MAR)}
\begin{figure}
\centering
\includegraphics[width=0.5\linewidth]{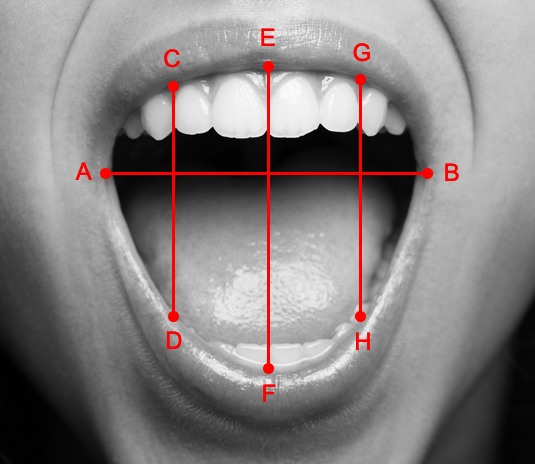}
\caption{Mouth aspect ratio}
\label{fig:mar}
\end{figure}

Computationally similar to the EAR, the MAR measures the ratio of the length of the mouth to the width of the mouth, as illustrated in the Figure~\ref{fig:mar} and Equation~\ref{eqn:MAR} Researchers have previously pointed out that as an individual becomes drowsy, they are likely to yawn and lose control over their mouth, making their MAR to be higher than usual in this state~\cite{singh2018driver}.

\begin{equation} 
\label{eqn:MAR}
MAR=\frac{\left | CD \right | + \left | EF \right | + \left | GH \right |}{3\ast \left | AB \right |}
\end{equation}

\subsubsection{Pupil Circularity (PUC)}
PUC is a measure complementary to EAR, but it places a greater emphasis on the pupil instead of the entire eye, as illustrated in Equations \ref{eqn:PUC1}, \ref{eqn:PUC2}, and \ref{eqn:PUC3}. For example, someone who has their eyes half-open or almost closed will have a much lower pupil circularity value versus someone who has their eyes fully open due to the squared term in the denominator.

\begin{equation} 
\label{eqn:PUC1}
Circularity=\frac{4\ast \pi \ast Area}{Perimeter^{2}}
\end{equation}

\begin{equation} 
\label{eqn:PUC2}
Area=\left ( \frac{Distance\left ( p_{2},p_{5} \right )}{2} \right )^{2}\ast \pi
\end{equation}

\begin{equation} 
\begin{aligned}
\label{eqn:PUC3}
Perimeter=Distance\left ( p_{1},p_{2} \right )+Distance\left ( p_{2},p_{3} \right )+Distance\left ( p_{3},p_{4} \right ) \\
+Distance\left ( p_{4},p_{5} \right )+Distance\left ( p_{5},p_{6} \right )+Distance\left ( p_{6},p_{1} \right )
\end{aligned}
\end{equation}

\subsubsection{Mouth Aspect Ratio Over Eye Aspect Ratio (MOE)}
MOE is the ratio of the MAR to the EAR. The benefit of using this feature is that EAR and MAR are expected to move in opposite directions if the state of the individual changes. As opposed to both EAR and MAR, MOE as a measure will be more responsive to these changes as it will capture the subtle changes in both EAR and MAR and will exaggerate the changes as the denominator and numerator move in opposite directions. As an individual gets drowsy, the MOE will increase.

\subsubsection{Feature Normalization}
We did feature normalization because each subject has different core features in their default alert state. That is, person A may naturally have smaller eyes than person B. If a model is trained on person B while tested on person A, the classifier will always predict the state as drowsy because it will detect a fall in EAR and PUC and a rise in MOE even though person A was alert.

To normalize the features of each training subject, we took the first 30 frames for each training subject’s alert video and used them as the baseline. The mean and standard deviation of each feature for these 30 frames were calculated and used to normalize each feature individually for each test subject. Mathematically, this is what the normalization equation looked like:
\begin{equation} 
Normalised Feature_{f,s}=\frac{Feature_{f,s}-Mean_{f,s}}{STD_{f,s}}
\end{equation}
where: $f$ is the feature, $s$ is the subject, mean and standard deviation are taken from the first 30 frames of the alert state video.

\subsection{Classification Algorithm}

\subsubsection{K Nearest Neighbors}
\begin{figure}
\centering
\includegraphics[width=\linewidth]{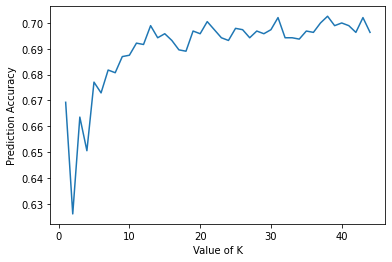}
\caption{Prediction accuracy on the test data set for K from 1 to 45}
\label{fig:acc}
\end{figure}

\begin{figure}
\centering
\includegraphics[width=\linewidth]{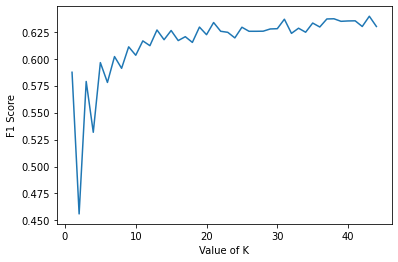}
\caption{F1 score on the test data set for K from 1 to 45}
\label{fig:f1s}
\end{figure}

Motivated by the article \cite{wang2019research, zhong_2019}, the KNN classifier obtained useful results. K-nearest neighbors with the kd-trees search algorithm were used for the classification task. We extracted features described in section \ref{sec:features} from videos in the UTA-RLDD data set. We used 80\% of the data as the training data set and 20\% of the data as the test data set. Based on the features and the corresponding label (alert or drowsy) in the training data set, we trained the K neighbors classifier to fit the model. We took 45 runs for K from 1 to 45 to find the best K value to predict the alertness level of the features in the test data set. Figure \ref{fig:acc} shows the prediction accuracy of the K neighbors classifier on the test data set with the different K values from 1 to 45. Figure \ref{fig:f1s} shows the F1 score of the K neighbors classifier on the test data set with the different K values from 1 to 10. We chose the K value as 38 to achieve the highest alertness levels prediction accuracy (>70\%) on the test data set.

\section{Results}
\subsection{Dataset and Preprocessing}
For our training and test data, we used the Real-Life Drowsiness Dataset (UTA-RLDD) \cite{ghoddoosian2019realistic} created specifically for detecting multi-stage drowsiness. The dataset consists of around 30 hours of videos of 60 unique participants. There were 51 men and 9 women, from different ethnicities (10 Caucasian, 5 non-white Hispanic, 30 Indo-Aryan and Dravidian, 8 Middle Eastern, and 7 East Asian) and ages (from 20 to 59 years old with a mean of 25 and standard deviation of 6). The subjects wore glasses in 21 of the 180 videos and had considerable facial hair in 72 out of the 180 videos. Videos were taken from roughly different angles in different real-life environments and backgrounds. Participants were instructed to take three videos of themselves by their phone/web camera (of any model or type) in three different drowsiness states, based on the Karolinska Sleep Scale~\cite{shahid2011karolinska}, for around 10 minutes each. All videos were recorded at such an angle that both eyes were visible, and the camera was placed within one arm length away from the subject. These instructions were used to make the videos similar to videos that would be obtained in a car, by phone placed in a phone holder on the dash of the car while driving. The proposed setup was to lay the phone against the display of their laptop while they were watching or reading something on their computer.

For each video, we used OpenCV to extract 1 frame per second starting at the 40 seconds mark until the end of the video. The end goal is to detect not only extreme and visible cases of drowsiness but allow our system to detect softer signals of drowsiness as well. 

\subsection{Results on the UTA-RLDD dataset}
In this section, we reported the accuracy, F1-score, precision, and recall of the models we trained using the K-nearest neighbors algorithm to predict the drowsy state when using different sets of features. Table~\ref{tab:nthures} shows the detailed information.

We extracted 13235 frames from the UTA-RLDD dataset. We used 9264 frames as the training data and the remaining 3971 frames as test data. Among the 9264 frames in the training dataset, 51\% were extracted from the alert state videos, and 49\% were extracted from the drowsy state videos. Among the 3971 frames in the test dataset, 45.5\% were extracted from the alert state videos, and 54.5\% were extracted from the drowsy state videos.

\begin{table*}[]
\begin{tabular}{|l|l|l|l|l|}
\hline
Features                 & Accuracy         & Precision     & Recall        & F1 Score      \\ \hline
EAR                      & 79.53\%          & 0.74          & 0.87          & 0.80          \\ \hline
MAR                      & 49.94\%          & 0.31          & 0.58          & 0.40          \\ \hline
PUC                      & 71.42\%          & 0.65          & 0.79          & 0.71          \\ \hline
\textbf{MOE}             & \textbf{85.04\%} & \textbf{0.79} & \textbf{0.93} & \textbf{0.85} \\ \hline
EAR+MAR                  & 47.87\%          & 0.049         & 0.91          & 0.093         \\ \hline
EAR+PUC                  & 60.59\%          & 0.46          & 0.72          & 0.56          \\ \hline
\textbf{EAR+MOE}         & \textbf{85.62\%} & \textbf{0.78} & \textbf{0.95} & \textbf{0.85} \\ \hline
MAR+PUC                  & 55.50\%          & 0.25          & 0.78          & 0.38          \\ \hline
\textbf{MAR+MOE}         & \textbf{86.33\%} & \textbf{0.79} & \textbf{0.95} & \textbf{0.86} \\ \hline
\textbf{PUC+MOE}         & \textbf{85.32\%} & \textbf{0.79} & \textbf{0.93} & \textbf{0.85} \\ \hline
EAR+MAR+PUC              & 49.23\%          & 0.16          & 0.64          & 0.25          \\ \hline
\textbf{EAR+MAR+MOE}     & \textbf{85.72\%} & \textbf{0.78} & \textbf{0.95} & \textbf{0.86} \\ \hline
\textbf{EAR+PUC+MOE}     & \textbf{84.56\%} & \textbf{0.76} & \textbf{0.94} & \textbf{0.84} \\ \hline
\textbf{MAR+PUC+MOE}     & \textbf{85.55\%} & \textbf{0.78} & \textbf{0.95} & \textbf{0.85} \\ \hline
\textbf{EAR+MAR+PUC+MOE} & \textbf{85.19\%} & \textbf{0.77} & \textbf{0.95} & \textbf{0.85} \\ \hline
\end{tabular}
\caption{Accuracy, precision, recall, F1 score when using different combinations of features.}
\label{tab:nthures}
\end{table*}

\begin{table*}[]
\begin{tabular}{ccccccccc}
\hline
\multicolumn{1}{|c|}{}       & \multicolumn{2}{c|}{EAR}                                & \multicolumn{2}{c|}{MAR}                               & \multicolumn{2}{c|}{PUC}                                & \multicolumn{2}{c|}{MOE}                              \\ \hline
\multicolumn{1}{|c|}{}       & \multicolumn{1}{c|}{Mean}   & \multicolumn{1}{c|}{Std}  & \multicolumn{1}{c|}{Mean}  & \multicolumn{1}{c|}{Std}  & \multicolumn{1}{c|}{Mean}   & \multicolumn{1}{c|}{Std}  & \multicolumn{1}{c|}{Mean} & \multicolumn{1}{c|}{Std}  \\ \hline
\multicolumn{1}{|c|}{Alert}  & \multicolumn{1}{c|}{0.28}   & \multicolumn{1}{c|}{0.06} & \multicolumn{1}{c|}{0.98}  & \multicolumn{1}{c|}{0.12} & \multicolumn{1}{c|}{0.43}   & \multicolumn{1}{c|}{0.06} & \multicolumn{1}{c|}{3.86} & \multicolumn{1}{c|}{1.62} \\ \hline
\multicolumn{1}{|c|}{Drowsy} & \multicolumn{1}{c|}{0.21}   & \multicolumn{1}{c|}{0.10} & \multicolumn{1}{c|}{1.04}  & \multicolumn{1}{c|}{0.13} & \multicolumn{1}{c|}{0.38}   & \multicolumn{1}{c|}{0.07} & \multicolumn{1}{c|}{7.12} & \multicolumn{1}{c|}{5.04} \\ \hline
\multicolumn{1}{|c|}{Delta}    & \multicolumn{1}{c|}{-25.0\%} & \multicolumn{1}{c|}{}      & \multicolumn{1}{c|}{+6.1\%} & \multicolumn{1}{c|}{}      & \multicolumn{1}{c|}{-16.7\%} & \multicolumn{1}{c|}{}      & \multicolumn{1}{c|}{+84.5} & \multicolumn{1}{c|}{}     \\ \hline
\end{tabular}
\caption{Mean and standard deviation of different feature for the alert and drowsy conditions}
\label{tab:nthumean}
\end{table*}

The result shows that MOE (the ratio of MAR divided by EAR) seems to be the most sensitive feature among the four features we tested. The result aligns with the fact that MOE captures the subtle changes in both EAR and MAR and will exaggerate the changes as the denominator and numerator move in opposite directions. Though the feature MAR is the least sensitive feature among the four features (accuracy=49.94\%, F1 score=0.4), the highest accuracy (86.33\%) and F1 score (0.86) were found when using the feature combination of MAR and MOE, followed by using the feature combination of EAR, MAR, and MOE (accuracy=85.72\%, F1 score=0.86). The third highest accuracy was found using the feature combination of EAR and MOE (accuracy=85.63\%, F1 score=0.85).

We also found that in some cases, individual features work better than a combination of features. For example, when using the feature combinations EAR+MAR, EAR+PUC, MAR+PUC, and EAR+MAR+PUC, the accuracy, and F1 score are lower than using these features individually. 

We investigated the mean and standard deviation of the EAR, MAR, PUC, and MOE for alert and drowsy conditions. Table~\ref{tab:nthumean} shows that EAR and PUC are larger in the alert condition than in the drowsy condition on average, while the MAR is smaller in the alert condition than in the drowsy state. The result also shows that the MOE is larger in the drowsy condition than in the alert condition on average. Moreover, the difference in MOE's mean value in the alert and the drowsy condition is larger than other than three features. This aligns with the fact that MOE could exaggerate the changes of EAR and MAR.

\subsection{Results on target population}
\begin{figure}
\centering
\includegraphics[width=\linewidth]{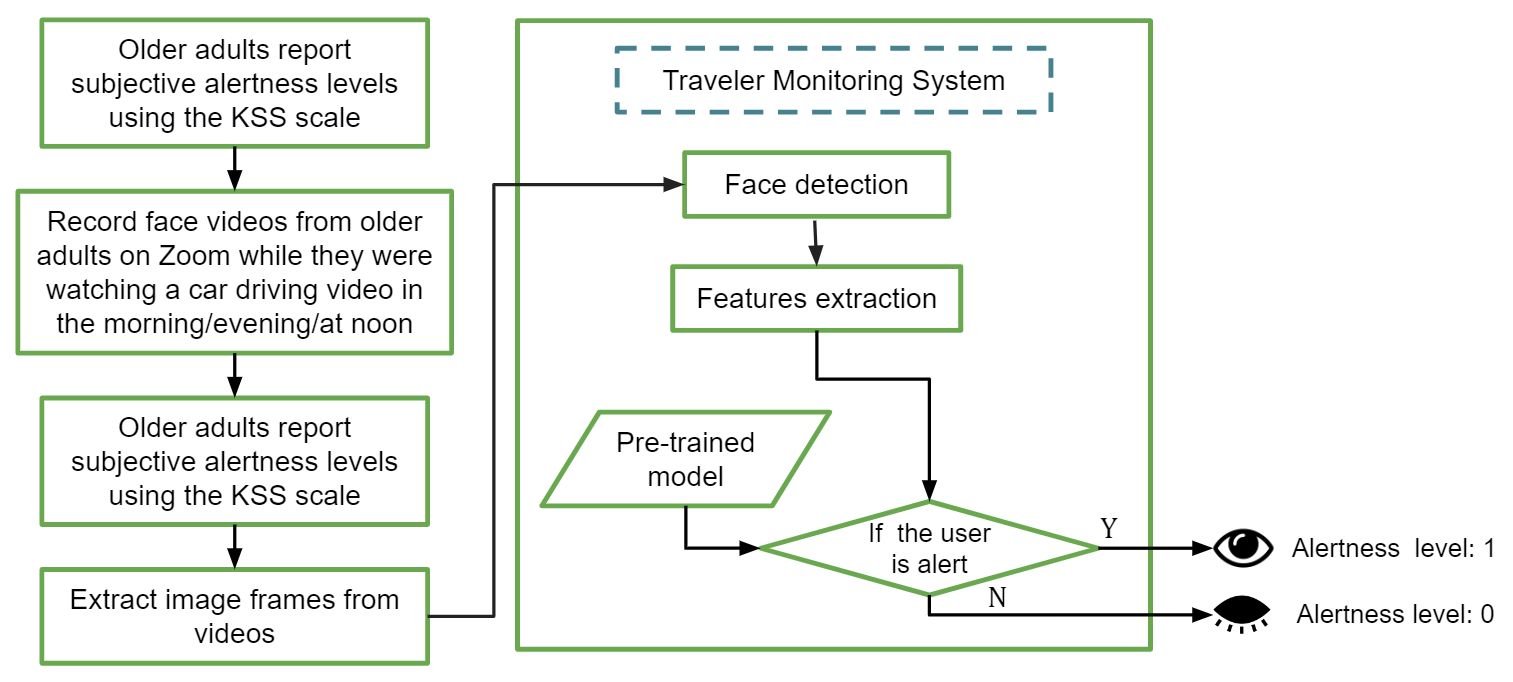}
\caption{Detailed study procedure}
\label{fig:studyprocedure}
\end{figure}

We tested our algorithm on older adult participants to evaluate how well we could detect drowsiness in older adults. Older adults were recruited from the general population. Ten participants agreed to participate in three Zoom sessions, one in the morning, one around noon, and one in the evening. These times were selected so as to record videos of participants' faces when they were alert and drowsy. In the Zoom meeting, participants were asked to watch a car driving video~\footnote{https://www.youtube.com/watch?v=fkps18H3SXY}. Participants were instructed to feel free to talk or look around as they usually do when traveling as a passenger. Participants were asked to report their sleepiness level based on the Karonlinksa Sleep Scale~\cite{shahid2011karolinska} before and after watching the car driving video. Figure~\ref{fig:studyprocedure} shows the detailed study procedure. Table \ref{tab:participantresults} shows the meeting time with participants and their self-reported sleepiness level.

We collected 33 videos from 10 older adults without MCI. Table \ref{tab:participantresults} shows the percentage of frames that the traveler monitoring system could detect the participant’s facial landmark. The results show that the facial landmark detection rate for most participants is higher than 90\% except for participants 7, 8, and 9. The main reasons that the results were low for these three participants are that (1) their video resolution was low and (2) their face was out of the viewport of the camera. To address these concerns, we used a higher-resolution Logitech c920 webcam in our final prototype system while placing the camera directly in front of the user to capture their face during the entire process; thus, ensuring our system can detect greater than 90\% of facial landmarks.

Participant 2 reported they were extremely alert in the first session, alert in the second session, very alert in the third session, and sleepy in the fourth session. Based on their reported subjective alertness level, their alertness level is highest in session A and lowest in session D. Table \ref{tab:participant2example} shows the mean and standard deviation of the EAR, MAR, PUC, and MOE for that participant. The average EAR and PUC are the highest, while the average MOE is the lowest in session A. The average MAR and MOE are the highest, while the average PUC is the lowest in session D. The bolded values in table \ref{tab:participant2example} show that the MOE features can distinguish the different alertness levels, and that MOE is a good feature to predict the alertness level.

\begin{table*}[tb]
\begin{tabular}{|c|c|cc|cc|cc|cc|}
\hline
\multirow{2}{*}{} & \multirow{2}{*}{Self-reported alertness} & \multicolumn{2}{c|}{Eye aspect ratio (EAR)} & \multicolumn{2}{c|}{Mouth aspect ratio (MAR)} & \multicolumn{2}{c|}{Pupil circularity (PUC)} & \multicolumn{2}{c|}{MAR/EAR (MOE)}          \\ \cline{3-10} 
                  &                                          & \multicolumn{1}{c|}{Mean}       & SD        & \multicolumn{1}{c|}{Mean}        & SD         & \multicolumn{1}{c|}{Mean}       & SD         & \multicolumn{1}{c|}{Mean}           & SD    \\ \hline
Session A         & \textbf{1}                               & \multicolumn{1}{c|}{0.289}      & 0.047     & \multicolumn{1}{c|}{0.991}       & 0.049      & \multicolumn{1}{c|}{0.438}      & 0.051      & \multicolumn{1}{c|}{\textbf{3.546}} & 0.784 \\ \hline
Session B         & 3                                        & \multicolumn{1}{c|}{0.264}      & 0.028     & \multicolumn{1}{c|}{1.024}       & 0.030      & \multicolumn{1}{c|}{0.393}      & 0.033      & \multicolumn{1}{c|}{3.934}          & 0.488 \\ \hline
Session C         & 2                                        & \multicolumn{1}{c|}{0.247}      & 0.036     & \multicolumn{1}{c|}{0.976}       & 0.043      & \multicolumn{1}{c|}{0.394}      & 0.038      & \multicolumn{1}{c|}{4.073}          & 0.838 \\ \hline
Session D         & \textbf{8}                               & \multicolumn{1}{c|}{0.257}      & 0.021     & \multicolumn{1}{c|}{1.174}       & 0.070      & \multicolumn{1}{c|}{0.393}      & 0.025      & \multicolumn{1}{c|}{\textbf{4.602}} & 0.496 \\ \hline
\end{tabular}
\caption{Mean and standard deviation of alertness detection features of participant 2 in four sessions}
\label{tab:participant2example}
\end{table*}

\begin{table*}[]
\begin{tabular}{|c|c|c|c|c|c|}
\hline
Participant ID & Session time  & \begin{tabular}[c]{@{}c@{}}Before video self-reported\\ alertness level\end{tabular} & \begin{tabular}[c]{@{}c@{}}After video self-reported\\ alertness level\end{tabular}               & \begin{tabular}[c]{@{}c@{}}After video self-reported\\ sleepiness symptoms\end{tabular} & \begin{tabular}[c]{@{}c@{}}Facial landmark\\ detection rate \%\end{tabular} \\ \hline
2A             & 12 PM - 1 PM  & Fairly alert                                                                         & Extremely alert                                                                                   & \begin{tabular}[c]{@{}c@{}}occurred during\\ 25\% of the time\end{tabular}              & 100                                                                         \\ \hline
2B             & 11 AM - 12 PM & \begin{tabular}[c]{@{}c@{}}Sleepy, but no \\ effort to keep alert\end{tabular}       & \begin{tabular}[c]{@{}c@{}}Some signs of \\ sleepiness\end{tabular}                               & \begin{tabular}[c]{@{}c@{}}occurred during\\ 50\% of the time\end{tabular}              & 100                                                                         \\ \hline
2C             & 6 PM - 7 PM   & Very alert                                                                           & Very alert                                                                                        & did not occur                                                                           & 99.1                                                                        \\ \hline
2D             & 9 PM - 10 PM  & \begin{tabular}[c]{@{}c@{}}Sleepy, but no \\ effort to keep alert\end{tabular}       & \begin{tabular}[c]{@{}c@{}}Sleepy, some effort \\ to keep alert\end{tabular}                      & \begin{tabular}[c]{@{}c@{}}occurred during\\ 50\% of the time\end{tabular}              & 100                                                                         \\ \hline
3A             & 8 AM - 9 AM   & \begin{tabular}[c]{@{}c@{}}Some signs of \\ sleepiness\end{tabular}                  & Very alert                                                                                        & did not occur                                                                           & 99.8                                                                        \\ \hline
3B             & 11 AM - 12 PM & \begin{tabular}[c]{@{}c@{}}Neither alert \\ nor sleepy\end{tabular}                  & Fairly alert                                                                                      & \begin{tabular}[c]{@{}c@{}}occurred a\\ few times\end{tabular}                          & 95.0                                                                        \\ \hline
3C             & 5 PM - 6 PM   & Very alert                                                                           & \begin{tabular}[c]{@{}c@{}}Some signs of \\ sleepiness\end{tabular}                               & \begin{tabular}[c]{@{}c@{}}occurred a\\ few times\end{tabular}                          & 99.9                                                                        \\ \hline
4A             & 10 AM - 11 AM & Very alert                                                                           & Very alert                                                                                        & did not occur                                                                           & 91.9                                                                        \\ \hline
4B             & 6 PM - 7 PM   & Very alert                                                                           & Alert                                                                                             & \begin{tabular}[c]{@{}c@{}}occurred a\\ few times\end{tabular}                          & 99.1                                                                        \\ \hline
4C             & 1 PM - 2 PM   & Very alert                                                                           & Very alert                                                                                        & did not occur                                                                           & 88.6                                                                        \\ \hline
5A             & 9 AM - 10 AM  & Alert                                                                                & Very alert                                                                                        & did not occur                                                                           & 99.5                                                                        \\ \hline
5B             & 1 PM - 2 PM   & Very alert                                                                           & Very alert                                                                                        & did not occur                                                                           & 99.5                                                                        \\ \hline
5C             & 6 PM - 7 PM   & Very alert                                                                           & Very alert                                                                                        & did not occur                                                                           & 99.9                                                                        \\ \hline
5D             & 9 PM - 10 PM  & Fairly alert                                                                         & \begin{tabular}[c]{@{}c@{}}Sleepy, some effort\\ to keep alert\end{tabular}                       & \begin{tabular}[c]{@{}c@{}}occurred during\\ 25\% of the time\end{tabular}              & 99.6                                                                        \\ \hline
6A             & 8 AM - 9 AM   & Alert                                                                                & Fairly alert                                                                                      & did not occur                                                                           & 100                                                                         \\ \hline
6B             & 7 PM - 8 PM   & \begin{tabular}[c]{@{}c@{}}Neither alert \\ nor sleepy\end{tabular}                  & \begin{tabular}[c]{@{}c@{}}Sleepy, some effort\\ to keep alert\end{tabular}                       & \begin{tabular}[c]{@{}c@{}}occurred during\\ 25\% of the time\end{tabular}              & 52.5                                                                        \\ \hline
6C             & 12 PM - 1 PM  & Very alert                                                                           & \begin{tabular}[c]{@{}c@{}}Some signs of\\ sleepiness\end{tabular}                                & \begin{tabular}[c]{@{}c@{}}occurred during\\ 50\% of the time\end{tabular}              & 100                                                                         \\ \hline
7A             & 8 AM - 9 AM   & Very alert                                                                           & Alert                                                                                             & did not occur                                                                           & 0                                                                           \\ \hline
7B             & 12 PM - 1 PM  & Alert                                                                                & Alert                                                                                             & did not occur                                                                           & 61.3                                                                        \\ \hline
7C             & 5 PM - 6 PM   & Alert                                                                                & Alert                                                                                             & did not occur                                                                           & 1.2                                                                         \\ \hline
8A             & 9 AM - 10 AM  & Alert                                                                                & \begin{tabular}[c]{@{}c@{}}Neither alert\\ nor sleepy\end{tabular}                                & \begin{tabular}[c]{@{}c@{}}occurred a\\ few times\end{tabular}                          & 70.4                                                                        \\ \hline
8B             & 1 PM - 2 PM   & Fairly alert                                                                         & \begin{tabular}[c]{@{}c@{}}Sleepy, some effort \\ to keep alert\end{tabular}                      & \begin{tabular}[c]{@{}c@{}}occurred during\\ 50\% of the time\end{tabular}              & 84.7                                                                        \\ \hline
8C             & 6 PM - 7 PM   & Fairly alert                                                                         & \begin{tabular}[c]{@{}c@{}}Sleepy, some effort\\ to keep alert\end{tabular}                       & \begin{tabular}[c]{@{}c@{}}occurred during\\ 25\% of the time\end{tabular}              & 2.4                                                                         \\ \hline
8D             & 2 PM - 3 PM   & \begin{tabular}[c]{@{}c@{}}Sleepy, some effort \\ to keep alert\end{tabular}         & \begin{tabular}[c]{@{}c@{}}Very sleepy, great effort\\ to keep alert, fighting sleep\end{tabular} & \begin{tabular}[c]{@{}c@{}}occurred most\\ of the time\end{tabular}                     & 17.8                                                                        \\ \hline
9A             & 9 AM - 10 AM  & \begin{tabular}[c]{@{}c@{}}Some signs of \\ sleepiness\end{tabular}                  & \begin{tabular}[c]{@{}c@{}}Some signs of\\ sleepiness\end{tabular}                                & \begin{tabular}[c]{@{}c@{}}occurred during\\ 25\% of the time\end{tabular}              & 0.3                                                                         \\ \hline
9B             & 2 PM - 3 PM   & \begin{tabular}[c]{@{}c@{}}Some signs of \\ sleepiness\end{tabular}                  & \begin{tabular}[c]{@{}c@{}}Some signs of\\ sleepiness\end{tabular}                                & \begin{tabular}[c]{@{}c@{}}occurred a\\ few times\end{tabular}                          & 9.6                                                                         \\ \hline
9C             & 9 PM - 10 PM  & \begin{tabular}[c]{@{}c@{}}Neither alert \\ nor sleepy\end{tabular}                  & \begin{tabular}[c]{@{}c@{}}Some signs of\\ sleepiness\end{tabular}                                & \begin{tabular}[c]{@{}c@{}}occurred a\\ few times\end{tabular}                          & 48.3                                                                        \\ \hline
10A            & 9 AM - 10 AM  & Very alert                                                                           & Alert                                                                                             & did not occur                                                                           & 92.4                                                                        \\ \hline
10B            & 1 PM - 2 PM   & Alert                                                                                & Fairly alert                                                                                      & \begin{tabular}[c]{@{}c@{}}occurred a\\ few times\end{tabular}                          & 86.6                                                                        \\ \hline
10C            & 6 PM - 7 PM   & Alert                                                                                & Alert                                                                                             & did not occur                                                                           & 87.6                                                                        \\ \hline
11A            & 4 PM - 5 PM   & Alert                                                                                & \begin{tabular}[c]{@{}c@{}}Neither alert\\ nor sleepy\end{tabular}                                & did not occur                                                                           & 96.8                                                                        \\ \hline
11B            & 9 AM - 10 AM  & Very alert                                                                           & Very alert                                                                                        & did not occur                                                                           & 99.7                                                                        \\ \hline
11C            & 3 PM - 4 PM   & Alert                                                                                & Alert                                                                                             & did not occur                                                                           & 99.8                                                                        \\ \hline
\end{tabular}
\caption{Meeting time with older adults and their subjective reported sleepiness level, and percentage of frames in the video that the traveler monitoring system can detect a facial landmark}
\label{tab:participantresults}
\end{table*} 
\section{Limitations and Future Work}
The first limitation of our system is that the outcome is bottlenecked by the success or failure of facial landmark detection. 

\begin{figure}[t]
\centering
\includegraphics[width=\linewidth]{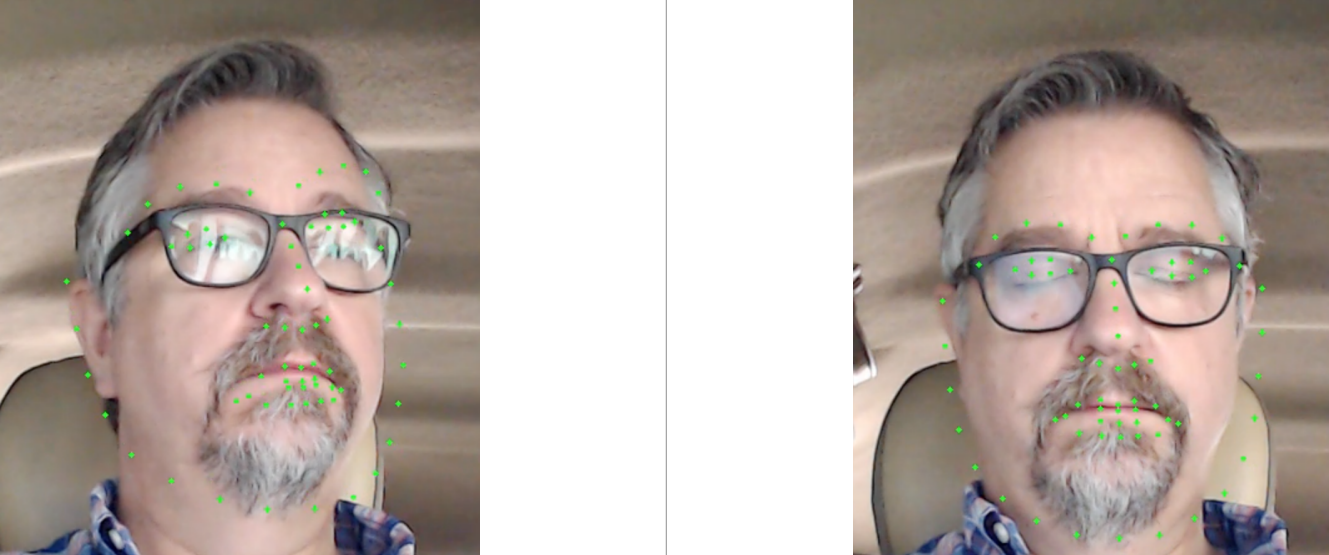}
\caption{Facial landmark false detection results when the user wearing glasses}
\label{fig:glasses}
\end{figure}

\textit{Glasses.} Almost all the older adults that participated in our study wore glasses. Wearing sunglasses also affects the performance of our system. Especially for our target population, it is very common to wear sunglasses because older adults need to protect their eyes from the sunlight. Facial landmark detection in the presence of glasses is a known challenge in the computer vision community. We found that eyeglasses affect the facial landmark detection because it is difficult to localize the eyes area due to light reflection and the frame of eyeglasses (Figure~\ref{fig:glasses} shows an example failure case). A potential solution for this issue is adding an image preprocessing step to remove glasses from the face images. Another possibility is to add a ``glasses detector'' as a preprocessing step and utilize cues such as head pose and head movement to detect sleepiness if glasses are detected. 

\textit{Lighting.} Facial landmark detection rate is also affected by the lighting in the scene. When vehicles run at night and the vehicle's interior becomes less visible, the performance of any visible light camera based system would be affected. A potential solution for this issue is applying dark image enhancement algorithms as a preprocessing step to facilitate facial landmarks detection in poor light conditions. Previously, researchers have proposed using near infrared lighting was used to address this issue~\cite{gwon2014gaze}. However, more than 50\% of sunlight is near infrared \cite{barolet2016infrared}. From a practical perspective, it might be necessary to have two types of camera in the vehicle, one for day use and one for night use.

\begin{figure}
\centering
\includegraphics[width=\linewidth]{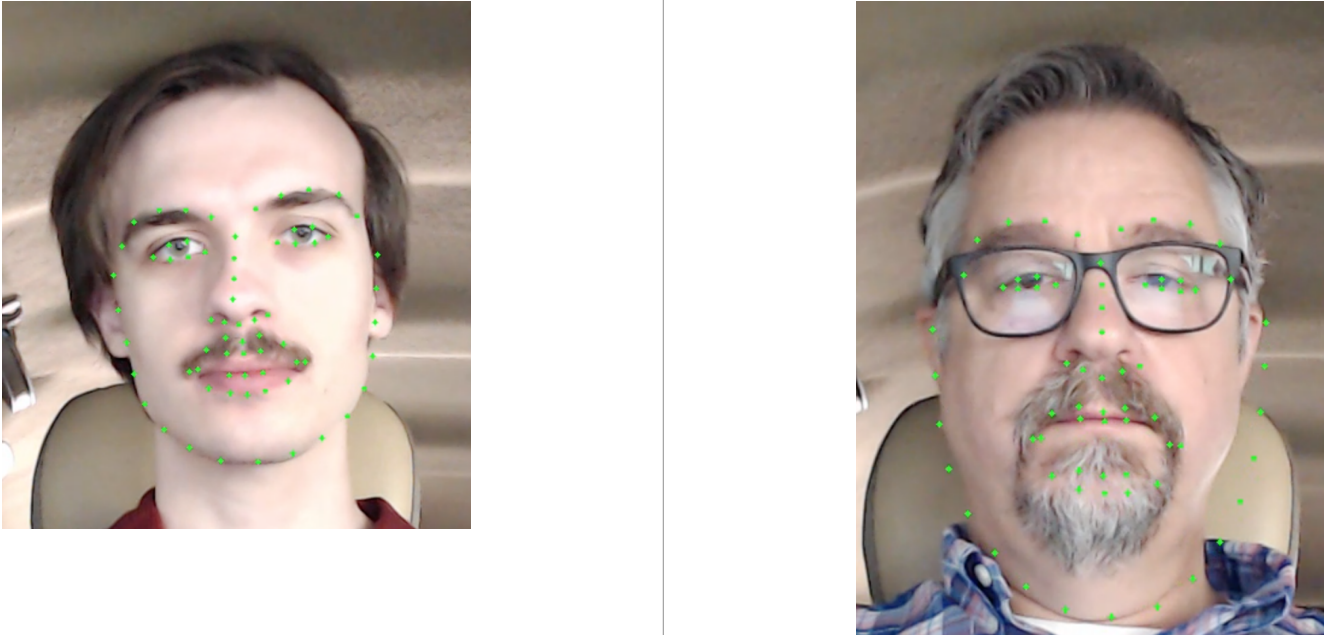}
\caption{Facial landmarks detection may be affected by facial hair.}
\label{fig:mustache}
\end{figure}

\begin{figure}[t]
\centering
\includegraphics[width=\linewidth]{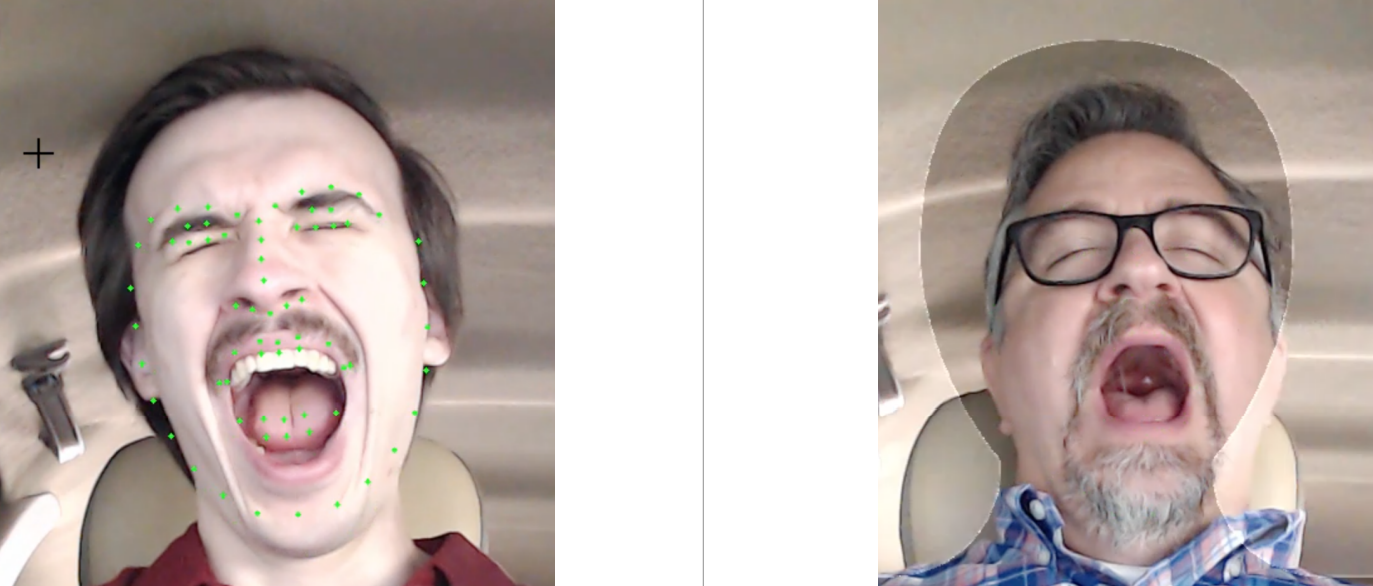}
\caption{Facial landmark detection results when users with facial hair are yawning.}
\label{fig:yawn}
\end{figure}

\textit{Facial hair.} The presence of facial hair is known to reduce the accuracy of facial landmarks detection. Figures~\ref{fig:mustache} and \ref{fig:yawn} illustrate how the subject's mustache and beard affected facial landmark detection. Future work might address these limitations by developing personalized facial landmarks detection algorithms, for example by asking the user to annotate a few pictures of themselves and using these annotations to learn the landmarks. In our use case, the vehicle is personally owned, and thus, we can expect the user to be the same at all times. 

\begin{figure}[t]
\centering
\includegraphics[width=0.5\linewidth]{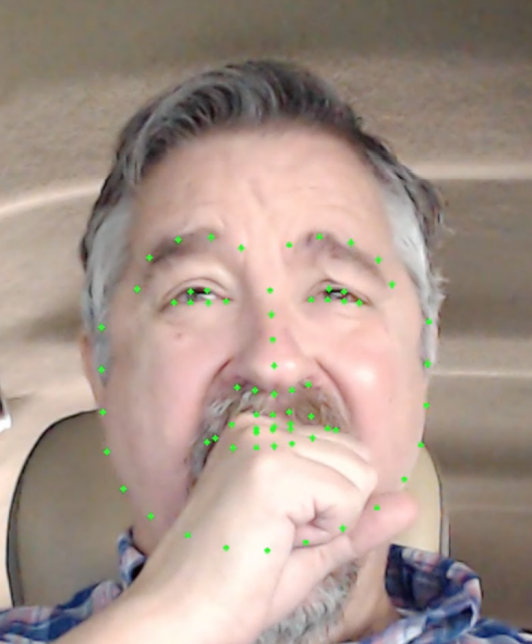}
\caption{Facial landmark detection is impacted when the user covers their mouth during a yawn.}
\label{fig:holdmouth}
\end{figure}

\textit{Behavioral differences.} The major limitation of our system, and possibly the richest area of future work, is accounting for how different people behave when they are inattentive, or drowsy or fatigued. For example, one of our test subjects put their hand over their mouth while yawning (Figure~\ref{fig:holdmouth}), which immediately led to false facial landmarks. Future work toward in-vehicle attentiveness monitoring in older adults should consider qualitative studies to elicit how the target population expresses themselves in the states that we need to detect algorithmically.

Moreover, the calibration stage significantly affect the detection accuracy. However, uses are not familiar about the whole system and don’t know what to do. We need to find a way to guarantee we can get accurate baseline features in the calibration stage. Users may need to stop detection in the trip or stop the alert to remind that the face is not at the center of the screen since they may need to move around. 

\bibliographystyle{ACM-Reference-Format}
\bibliography{mainbibliography}


\end{document}